\documentclass[letterpaper, 10 pt, conference]{ieeeconf}
\IEEEoverridecommandlockouts
\usepackage{cite}
\usepackage{amsmath,amssymb,amsfonts}
\usepackage{algorithm}
\usepackage{algpseudocode}
\usepackage{graphicx}
\usepackage{caption}
\usepackage{booktabs}
\usepackage{threeparttable}
\usepackage{multirow}
\usepackage{array}
\usepackage{listings}
\usepackage[most]{tcolorbox}
\usepackage{comment}
\usepackage{url}
\usepackage[hidelinks]{hyperref}
\usepackage{textcomp}
\usepackage{xcolor}
\usepackage[nolist]{acronym}
\usepackage{tikz}
\usepackage{colortbl}
\usepackage{siunitx}
\usepackage{makecell}
\usepackage[acronym, nomain]{glossaries}
\newacronym{vlm}{VLM}{ Vision Language Model}
\newacronym{vqa}{VQA}{Visual Question Answering}
\newacronym{rag}{RAG}{Retrieval-augmented Generation}
\newacronym{llm}{LLM}{Large Language Model}
\newacronym{mllm}{MLLM}{Multimodal Large Language Model}
\newacronym{ad}{AD}{Autonomous Driving}
\newacronym{odd}{ODD}{Operational Design Domain}
\newacronym{ads}{ADS}{Automated Driving System}
\newacronym{dsl}{DSL}{Domain Specific Language}
\newacronym{sotif}{SOTIF}{Safety of the Intended Functionality}
\newacronym{nl}{NL}{Natural Language}
\newacronym{bm25}{BM25}{Best Match 25}
\newacronym{rrf}{RRF}{Reciprocal Rank Fusion}
\newacronym{cp}{CP}{Contextual Prompting}
\newacronym{cot}{CoT}{Chain of Thought}
\newacronym{icl}{ICL}{In-Context Learning}
\newacronym{ragicl}{RAG-ICL}{Retrieval-Augmented In-Context Learning}
\newacronym{csr}{CSR}{Compilation Success Rate}
\newacronym{rt}{RT}{Response Time}
\newacronym{tokens}{Tokens}{Token Usage}
\newacronym{codeicl}{CodeICL}{Code In-Context Learning}
\newacronym{docicl}{DocICL}{Documentation In-Context Learning}
\newacronym{sq}{SQ}{Scenario Quality}
\newacronym{fa}{FA}{Framework Accuracy}
\newacronym{sota}{SOTA}{State-of-the-Art}
\usetikzlibrary{shapes.geometric, arrows, positioning}
\glsdisablehyper
\tikzstyle{block} = [rectangle, rounded corners, minimum width=3cm, minimum height=1cm,text centered, draw=black, fill=blue!20]
\tikzstyle{input} = [ellipse, minimum width=2cm, minimum height=1cm, text centered, draw=black, fill=green!20]
\tikzstyle{output} = [ellipse, minimum width=2cm, minimum height=1cm, text centered, draw=black, fill=red!20]
\tikzstyle{arrow} = [thick,->,>=stealth]

\graphicspath{{figures/}}

\def\BibTeX{{\rm B\kern-.05em{\sc i\kern-.025em b}\kern-.08em
    T\kern-.1667em\lower.7ex\hbox{E}\kern-.125emX}}

\makeatletter
\def\ps@firstpagefooter{%
  \def\@oddhead{}\def\@evenhead{}%
  \def\@oddfoot{\hss\fbox{\parbox{\dimexpr\textwidth-2\fboxsep-2\fboxrule\relax}{\footnotesize This work has been accepted to the IEEE IROS 2026. Copyright may be transferred without notice, after which this version may no longer be accessible.}}\hss}%
  \def\@evenfoot{}%
}
\makeatother

\makeatletter
\def\bstctlcite{\@ifnextchar[{\@bstctlcite}{\@bstctlcite[bcite]}}
\def\@bstctlcite[#1]#2{\@bsphack
  \@for\@citeb:=#2\do{%
    \edef\@citeb{\expandafter\@firstofone\@citeb}%
    \if@filesw\immediate\write\@auxout{\string\citation{\@citeb}}\fi}%
  \@esphack}
\makeatother

\begin{document}

\title{Chat2Scenic: An Iterative RAG-Based Framework for Scenario Generation in Autonomous Driving}
\author{
Yuan Gao$^{1,*}$,
Wenting Miao$^{1,*}$,
Mattia Piccinini$^{1}$,
Haoyu Wang$^{1}$,
Qunying Song$^{2}$,
Johannes Betz$^{1}$
\thanks{$^{1}$ Y. Gao, W. Miao, M. Piccinini, H. Wang and J. Betz are with the Professorship of Autonomous Vehicle Systems, TUM School of Engineering and Design, Technical University of Munich, 85748 Garching, Germany; Munich Institute of Robotics and Machine Intelligence (MIRMI)}
\thanks{$^{2}$ Q. Song is with University College London, London, United Kingdom}
\thanks{$^{*}$ Equal contribution.}
}

\maketitle
\thispagestyle{firstpagefooter}
\begin{abstract}
       Validating autonomous driving systems requires diverse, regulation-compliant test scenarios. In simulation-based testing, scenarios are defined as executable scripts. Yet automatically generating such scripts from regulatory descriptions remains an open challenge, and existing approaches face fundamental trade-offs. Retrieval-assemble methods achieve reasonable compilation rates but lack scalability, whereas retrieval-based full-script generation suffers from low compilation success rates. We present Chat2Scenic, the first iterative retrieval-augmented framework to generate scenario scripts in \gls{dsl}. Specifically, Chat2Scenic provides a chatbot interface that supports interactive scenario refinement and integrates \gls{rag} to ground scenario generation in regulatory knowledge and \gls{dsl} syntax. Furthermore, we propose an open benchmark for scenario generation comprising 123 scenarios from various regulations, including NHTSA and United Nations Vehicle Regulations, as well as other sources. Extensive evaluation with \gls{sota} \glspl{llm} demonstrates that Chat2Scenic achieves 76.42\% \gls{csr} and 58.17\% \gls{fa}, outperforming existing methods (Retrieval Assemble with 30.08\% \gls{csr}, 11.03\% \gls{fa} and Retrieval full script generation with 16.26\% \gls{csr}, 10.86\% \gls{fa}). To facilitate future research, we release our code as open source at \url{https://github.com/TUM-AVS/chat2scenic}.
\end{abstract}

\section{Introduction}
\label{sec:introduction}
Autonomous driving has matured significantly, with companies like Waymo deploying SAE Level 4 robotaxis~\cite{Betz2024} in urban environments, demonstrating fully-autonomous capabilities within specific \gls{odd}. This progress requires developing and validating highly reliable \gls{ads}. Traditional validation approaches depend on real-world testing, such as on-road trials~\cite{bäumler2024generating}. However, physical testing alone cannot adequately capture the diversity of driving situations and edge cases. To overcome these limitations, researchers and industry have increasingly adopted virtual scenario-based testing, which enables cost-effective simulation of diverse and realistic driving conditions~\cite{song2023critical}.  Recent advances in Large Language Models (LLMs) offer such potential: by leveraging knowledge of Domain Specific Language (DSL) syntax, they can automatically translate natural language scenario descriptions into executable scripts in \gls{dsl}.

\begin{figure}[t]
    \centering
    \includegraphics[width=0.95\linewidth]{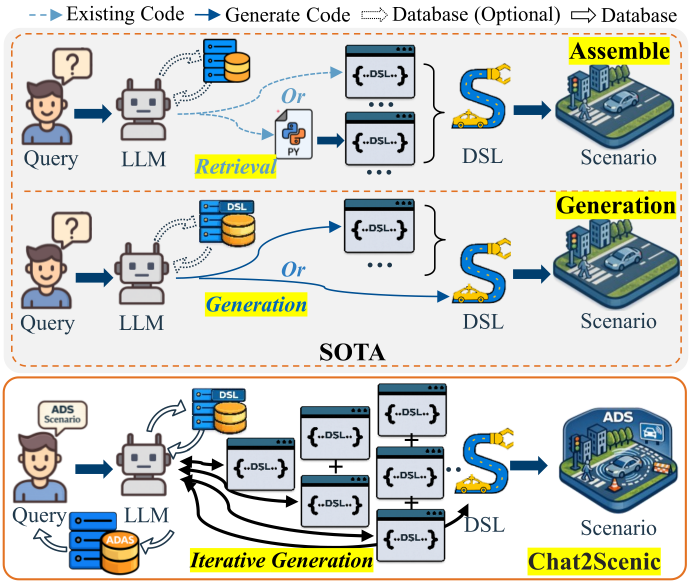}
    \vspace{-2mm}
    \caption{Comparison of \gls{dsl}-based scenario generation approaches and Chat2Scenic. While prior methods assemble existing DSL components or directly generate DSL programs, Chat2Scenic iteratively generates DSL blocks that interact with the database to create complex driving scenarios.}
    \label{fig:overview}
\end{figure}

Despite this progress, \gls{llm}-based generation of diverse and regulation-grounded scenarios remains challenging. As shown in Fig.~\ref{fig:overview}, existing automated approaches primarily rely on simplified scenario descriptions rather than complex \gls{ads} testing regulatory specifications. Moreover, current methods face fundamental trade-offs: retrieval-assemble approaches~\cite{deng2023target, zhang2024chatscene, cai2026text2scenario} achieve reasonable compilation rates by assembling pre-existing \gls{dsl} code snippets but lack generalizability to new scenarios, while direct generation~\cite{tian2024lmm, tang2024legend, elmaaroufi2024scenicnl, ruan2024traffic, rubavicius2024conversational, bauerfeind2025david} offers better flexibility but suffers from low compilation success rates. This motivates us to develop a framework that can handle complex \gls{ads} testing regulatory descriptions while maintaining both high compilation success and generalizability through iterative component-wise generation.

\subsection{Related Work}
\subsubsection{Classical Scenario Generation}
Many studies have explored the generation of \gls{ads} testing scenarios for scenario-based testing. 
One impactful survey~\cite{ding2023survey} categorizes generation methods into data-driven, adversarial, and knowledge-based approaches. In more detail, data-driven methods leverage real-world datasets to generate realistic scenarios. Adversarial approaches aim to expose \gls{ads} weaknesses by synthesizing rare but high-risk situations through techniques. Knowledge-based methods integrate domain expertise, such as traffic rules, to enhance controllability. Aligned with the ISO/WD PAS 21448 \gls{sotif} standard, Schutt et al.~\cite{schutt20231001} examine scenario generation across functional, logical, and concrete abstraction levels, covering machine learning-based generation, optimization-driven exploration, scenario extraction from driving data, and manual design.
\begin{table}[t]
\centering
\caption{Comparison of LLM-based Scenario Generation with \gls{dsl} format}
\label{tab:sota-comparison}
\begin{threeparttable}
\resizebox{\columnwidth}{!}{%
\renewcommand{\arraystretch}{1.1}
\setlength{\tabcolsep}{3pt}
\begin{tabular}{llccccccccc}
\toprule
& \multirow{2}{*}{\textbf{Paper}} & \multirow{2}{*}{\textbf{UI}} & \multicolumn{2}{c}{\textbf{RAG}} & \multicolumn{4}{c}{\textbf{Prompting Techniques}} & \multirow{2}{*}{\textbf{Reg.}} & \textbf{G} \\
\cmidrule(lr){4-5} \cmidrule(lr){6-9}
& & & \textbf{C} & \textbf{D} & \textbf{CP} & \textbf{CoT} & \textbf{ICL} & \textbf{RAG-ICL} & & \\
\midrule
\multirow{3}{*}{\rotatebox{90}{\textbf{Assemble}}} 
& 2023 TARGET~\cite{deng2023target} & $\times$ & $\times$ & $\times$ & \checkmark & $\times$ & \checkmark & $\times$ & $\times$ & C \\
& 2024 ChatScene~\cite{zhang2024chatscene} & $\times$ & \checkmark & $\times$ & $\times$ & $\times$ & \checkmark & $\times$ & $\times$ & C \\
& 2025 Text2Scenario~\cite{cai2026text2scenario} & $\times$ & $\times$ & $\times$ & $\times$ & \checkmark & \checkmark & $\times$ & $\times$ & C \\
\midrule
\multirow{8}{*}{\rotatebox{90}{\textbf{Generation}}} 
& 2024 LEADE~\cite{tian2024lmm} & $\times$ & $\times$ & $\times$ & $\times$ & \checkmark & \checkmark & $\times$ & $\times$ & F \\
& 2024 LeGEND~\cite{tang2024legend} & $\times$ & $\times$ & $\times$ & $\times$ & $\times$ & \checkmark & $\times$ & $\times$ & F \\
& 2024 ScenicNL~\cite{elmaaroufi2024scenicnl} & $\times$ & $\times$ & $\times$ & \checkmark & \checkmark & \checkmark & $\times$ & $\times$ & I \\
& 2024 Talk2Traffic~\cite{ruan2024traffic} & \checkmark & \checkmark & \checkmark & $\times$ & \checkmark & \checkmark & $\times$ & $\times$ & C \\
& 2025 Rubavicius~\cite{rubavicius2024conversational} & \checkmark & \checkmark & $\times$ & $\times$ & $\times$ & \checkmark  & \checkmark & $\times$ & F \\
& 2025 NL2Scenic~\cite{bauerfeind2025david} & $\times$ & \checkmark & $\times$ & \checkmark & \checkmark & \checkmark & \checkmark & $\times$ & F \\
 \rowcolor{gray!20} \cellcolor{white} & \textbf{Ours} & \checkmark & \checkmark & \checkmark & \checkmark & \checkmark & \checkmark & \checkmark & \checkmark & \textbf{I} \\
\bottomrule
\end{tabular}
}
\vspace{0.1em}
\begin{flushleft}
\scriptsize
UI: User Interaction. C: Code. D: Documentation RAG. CP: Contextual Prompting. CoT: Chain of Thought. ICL: In-Context Learning. RAG-ICL: Retrieval-Augmented In-Context Learning. Reg.: Regulatory Test Specifications. 
G - Granularity:  F (Full \gls{dsl}), C (Component-wise), I (Iterative Component-wise).
\end{flushleft}
\end{threeparttable}
\end{table}
\subsubsection{LLM-based Scenario Generation}
\glspl{llm} have emerged as powerful tools for generating scenarios directly from natural language inputs~\cite{gao2025foundation}.  Depending on the application, the underlying simulators fall into two categories: vehicle-centric simulators (e.g., CARLA~\cite{dosovitskiy2017carla}) and traffic-centric simulators (e.g., SUMO~\cite{SUMO2018}) for large-scale traffic generation. 
For traffic scenario generation, ChatSUMO~\cite{li2024chatsumo} and OmniTester~\cite{lu2024multimodal} leveraged \glspl{llm} to produce realistic urban traffic simulations from text, showcasing the scalability of language-driven scenario generation.

Recent survey~\cite{gao2025foundation} shows that \glspl{llm} have been used to generate scenarios in CARLA for autonomous driving based on \gls{dsl} formats like Scenic~\cite{elmaaroufi2024scenicnl, zhang2024chatscene, ruan2024traffic, rubavicius2024conversational, bauerfeind2025david} or OpenScenario~\cite{deng2023target, cai2026text2scenario}, or using simulator APIs~\cite{aiersilan2025generating, guo2024sovar, shi2025linguasim, lu2024realistic, zhou2024automatic}. However, API methods depend on large databases and mapping rules that bind \glspl{llm}-extracted elements to predefined templates, limiting expressiveness. \Glspl{dsl} provide a formal and human-readable representation for parameterized driving scenarios, corresponding to logical-level specifications that can be instantiated into concrete simulation setups. These \gls{dsl}-based methods are typically divided into two categories: Retrieval Assemble and Direct Generation, as shown in Table~\ref{tab:sota-comparison}. Retrieval Assemble retrieves the code snippets from the database or other code repositories and assembles them into a complete \gls{dsl} script. Direct Generation generates the \gls{dsl} script from scratch with or without retrieval.

\textbf{Retrieval Assemble:}
TARGET~\cite{deng2023target} parses traffic rules into logical scenarios via multi-stage prompting, then assembles concrete scripts through dictionary lookup and hierarchical map-based route retrieval. 
ChatScene~\cite{zhang2024chatscene} retrieves and assembles code snippets mapping behaviors to Scenic implementations using \gls{llm} with Retrieval-augmented Generation (RAG). Text2Scenario~\cite{cai2026text2scenario} performs \gls{llm}-based hierarchical scenario parsing with subsequent priority-based \gls{dsl} assembly to generate executable test cases. However, these retrieval-assemble methods lack generalizability to novel scenarios beyond their predefined code databases. 

\textbf{Direct Generation:}
LEADE~\cite{tian2024lmm} employs multi-modal prompting to generate executable scenario \gls{dsl} from abstract representations completely, followed by dual-layer optimization search for safety violations.
LeGEND~\cite{tang2024legend} uses two-stage \gls{llm} processing: translating reports into parameter patterns inserted into predefined templates, then generating concrete scripts from logical specifications. 
ScenicNL~\cite{elmaaroufi2024scenicnl} chains multiple \glspl{llm} with constrained decoding and uses compiler feedback for iterative refinement.
Talk2Traffic~\cite{ruan2024traffic} performs \gls{llm}-based retrieval-augmented snippet generation independently and followed by \gls{llm}-based component assembly. 
Rubavicius~\cite{rubavicius2024conversational} employs retrieval-augmented direct code generation with multi-turn dialogue refinement for Scenic-based autonomous driving scenario synthesis.
NL2Scenic~\cite{bauerfeind2025david} adopts full-program LLM-based Scenic generation with optional retrieval-augmented prompting. However, direct generation approaches suffer from low compilation rates due to the difficulty of generating complete scripts or assembling newly generated components, which are individually valid but mutually incompatible. 

\subsection{Critical Summary}
To the best of our knowledge, the existing literature is limited by at least one of the following aspects:
\begin{enumerate}
	\item Limited generative paradigms: Retrieval-based assembly methods~\cite{zhang2024chatscene, deng2023target, cai2026text2scenario} construct \gls{dsl} programs from stored code snippets, reducing generalization and scenario diversity.
	\item Compilation reliability and evaluation gaps: \gls{sota} methods focusing on direct \gls{dsl} generation via advanced prompting techniques~\cite{tian2024lmm, elmaaroufi2024scenicnl, ruan2024traffic, rubavicius2024conversational, bauerfeind2025david} face compilation reliability challenges, as \glspl{llm} generate either full scripts or components to assemble. Moreover, these approaches lack a comprehensive evaluation of compilation success rates and semantic accuracy.
	\item Limited input complexity: Most frameworks focus on simplified scenario descriptions rather than regulation-grounded specifications, which are essential for \gls{ads} validation in real-world deployment.
\end{enumerate}
\begin{figure*}[t]
	\centering
	\includegraphics[width=0.95\linewidth]{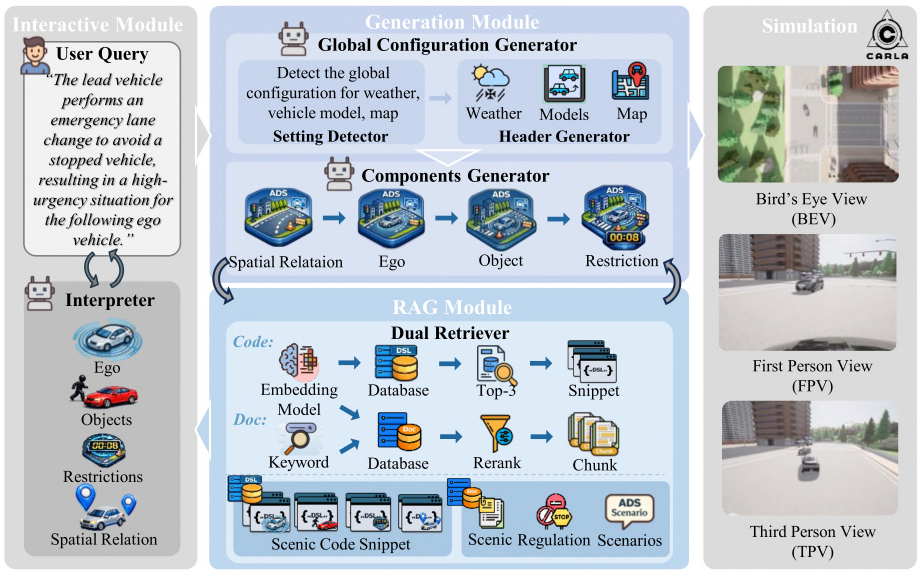}
    \vspace{-2mm}
	\caption{Overview of the Chat2Scenic framework, consisting of three core modules: the \textbf{Interactive Module} interactively refines and interprets user queries into logical components; the \textbf{Generation Module} synthesizes related \gls{dsl} snippets based on these components, integrating the \textbf{RAG Module} to retrieve domain specific knowledge.}
	\label{fig:framework}
\end{figure*}
\subsection{Contribution}
The key contributions of this paper are as follows:
\begin{enumerate}
	\item We propose \textbf{Chat2Scenic}, a chatbot-based iterative component-wise generation framework with advanced prompting techniques and \gls{rag}, outperforming State-of-the-Art (SOTA) methods with publicly available code across compilation success rate, semantic accuracy, and framework accuracy.
	\item We curate 120+ \gls{ads} scenario descriptions from the CARLA Leaderboard, NHTSA, and United Nations Vehicle Regulations, and propose dedicated metrics to evaluate both framework-level performance (e.g., compilation success rate and framework accuracy) and application-level performance (semantic accuracy).
	\item We evaluate Chat2Scenic across closed- and open-source \glspl{llm}, conduct ablation studies on prompting techniques, and benchmark against \gls{sota} methods.
\end{enumerate}

\section{Methodology}\label{sec:method}
Chat2Scenic consists of three parts (Fig.~\ref{fig:framework}): an interactive module, a \gls{rag} module, and a generation module, followed by a simulation interface. The interactive module parses the input description into a logical structured representation and supports interactive refinement. The \gls{rag} module retrieves domain-specific knowledge from a \gls{dsl} snippet database and a documentation database. The generation module iteratively generates Scenic code components until a complete Scenic program is produced, which is compatible with 3D simulators such as CARLA~\cite{dosovitskiy2017carla}. 

\subsection{Interactive Module}
Chat2Scenic provides an interactive framework through a chatbot interface supported by Gradio\footnote{\url{https://www.gradio.app/}}, as illustrated in Fig.~\ref{fig:chatbot}. Users can describe scenarios in natural language, query regulation-grounded scenarios retrieved from \gls{rag} databases, and iteratively refine their description. 

\subsubsection{Logical Structure Schema}
\begin{figure}[t]
	\centering
	\includegraphics[width=1\linewidth]{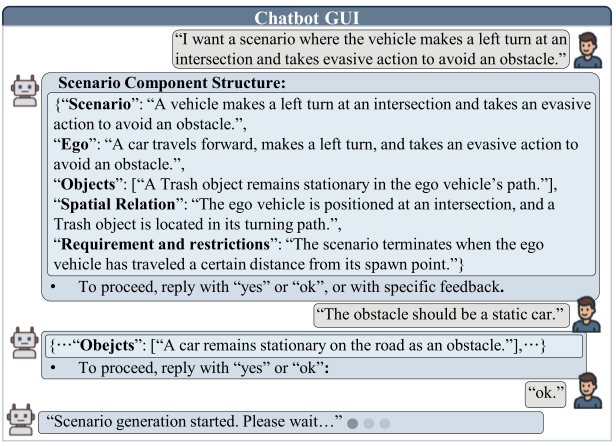}
    \vspace{-6mm}
	\caption{Chatbot UI for interactive refinement.}
	\label{fig:chatbot}
\end{figure}

To enable controllable, modular scenario generation, we represent each user request using a logical structure $\mathcal{S}$ by identifying the building blocks of the DSL. Specifically, we define the structured scenario representation $\mathcal{S}$ (Fig.~\ref{fig:chatbot}) as a set of components composed of global configuration $G$ and scenario components $\mathcal{S}_{int}$:
\begin{equation}
\mathcal{S} = \{G\} \cup \mathcal{S}_{int} = \{G, R, E, O, T\}
\end{equation}
Here, $G$  represents the global configuration (map, weather, vehicle models) and is independently extracted by the \textit{Global Configuration Generator} (Section~\ref{sec:global_config}). The \textit{Interpreter} extracts the scenario-specific components $\mathcal{S}_{int} = \{R, E, O, T\}$, where $R$ represents spatial relations (e.g., road topology and relative positioning of entities), $E$ denotes ego behavior (e.g., ego parameters), $O = \{O_1, ..., O_N\}$ is the set of $N$ objects (e.g., adversarial vehicles), and $T$ specifies restrictions (e.g., initial conditions and termination criteria).

\subsubsection{Interpreter} 
We employ an \gls{llm}-based interpreter in the workflow. The interpreter is prompted using explicit instructions and few-shot examples to extract $\mathcal{S}_{int}$ from scenario descriptions. Each component is abstracted into a single-sentence description, which is later used as a semantic query for component code generation. This abstraction aligns the interpretation with the structure of code snippets in the database and the Scenic syntax, facilitating effective retrieval and generation. The interpreter uses LangGraph's checkpointing mechanism\footnote{\url{https://www.langchain.com/langgraph}} with MemorySaver to maintain conversation state, enabling iterative refinement where users can provide feedback to update the logical interpretation without restarting the workflow.

\subsection{RAG Module}\label{subsec:rag}
To enhance the \gls{llm} performance for domain-specific tasks, we integrate \gls{rag} supported by the LangChain\footnote{\url{https://www.langchain.com/}} framework. The \gls{rag} technique retrieves relevant external knowledge and returns it to the \gls{llm} during scenario generation, improving accuracy and reducing hallucinations. Specifically, we construct two complementary databases: a code snippet database and a documentation database, and design specialized retrievers for each. 
\subsubsection{Code Snippet Database}
We construct a component-level database by collecting codes from official Scenic sources~\cite{fremont2019scenic}. Each scenario is decomposed into independent component-level code units covering the $\mathcal{S}_{int}$ components ($R$, $E$, $O$, $T$). For each component, an \gls{llm} generates a concise single-sentence description (e.g., ``A car travels forward and makes a right turn''). Each entry stores both the natural language description and the corresponding Scenic code snippet. The description is embedded using sentence transformers (all-MiniLM-L6-v2\footnote{\url{https://huggingface.co/sentence-transformers/all-MiniLM-L6-v2}}) for semantic retrieval, while the code snippet is returned as context to the generator.
\subsubsection{Documentation Database}
We construct a documentation database from official Scenic documentation and regulatory documents (e.g., UN Vehicle Regulations). Documents are collected via web crawling, cleaned to remove irrelevant content, and split into chunks using structure-aware hierarchical separators to preserve document hierarchy. Chunks are then embedded using an embedding model for semantic matching and indexed with \gls{bm25} for keyword-based retrieval, enabling a hybrid search that combines both approaches.

\subsubsection{Dual Retriever Architecture}
We employ a dual-retriever architecture to provide both code-level and language-level domain-specific context during generation. The \textbf{code retriever} queries the code snippet database using pure semantic search: given a component description from $\mathcal{S}_{int}$ (e.g., $E$: ``Ego vehicle travels forward''), it retrieves the top-3 most similar snippets filtered by component type via COSINE similarity. The \textbf{documentation retriever} uses a hybrid retrieval approach. For the same component description (e.g., $E$: ``Ego vehicle travels forward''), \gls{bm25} matches keywords and retrieves chunks containing explicit Scenic API definitions such as \texttt{FollowLaneBehavior}, while the embedding model retrieves semantically related content (e.g., general lane-following behavior descriptions). \gls{rrf} then merges and re-ranks both lists, so the returned chunks cover both exact API definitions and broader semantic background. This dual approach ensures the generator receives both concrete code patterns and domain-specific language knowledge.
\begin{algorithm}[t]
	\footnotesize
	\caption{Generation Pipeline}
	\label{alg:component_generator}
	\begin{algorithmic}[1]
	\State \textbf{Input:} $user\_query$ (Scenario description)
	\State \textbf{Output:} $\mathcal{S}_{code}$ (Scenic program)
	\State \textbf{--- Step 1: Logical Structure Extraction ---}
	\State $\mathcal{S}_{int} \gets \text{Interpreter}(user\_query)$ \Comment{extract $\{R, E, O, T\}$}
	\State \textbf{--- Step 2: Global Configuration ---}
	\State $(detected, confidence) \gets \text{SettingsDetector}(user\_query)$ 
	\For{$p \in \{map, weather, models\}$}
		\If{$confidence[p] \geq 0.6$ \textbf{and} $detected[p] \neq null$}
			\State $G_{settings}[p] \gets detected[p]$ \Comment{use detected value}
		\Else
			\State $G_{settings}[p] \gets default[p]$ \Comment{fall back to default}
		\EndIf
	\EndFor
	\State $G_{code} \gets \text{HeaderGenerator}(G_{settings})$ \Comment{Scenic header}
	\State \textbf{--- Step 3: Iterative Component Generation ---}
	\State $R_{code} \gets \text{Generator}_R(R, G_{code})$
	\State $E_{code} \gets \text{Generator}_E(E, G_{code}, R_{code})$ 
	\For{$i = 1$ to $N$} 
	    \State $O_{i,code} \gets \text{Generator}_O(O_i, G_{code}, R_{code}, E_{code}, O_{1:i-1,code})$ 
	\EndFor
	\State $T_{code} \gets \text{Generator}_T(T, G_{code}, R_{code}, E_{code}, O_{1:N,code})$ 
	\State $\mathcal{S}_{code} \gets \{G_{code}, R_{code}, E_{code}, O_{1:N,code}, T_{code}\}$
	\State \Return $\mathcal{S}_{code}$ 
	\end{algorithmic}
	\end{algorithm} 
\subsection{Generation Module}
The generation module transforms the logical representation $\mathcal{S}$ into executable Scenic code $\mathcal{S}_{code}$ as shown in Algorithm~\ref{alg:component_generator}. The process consists of two main stages: (1) global configuration generation and (2) iterative component-wise generation with context accumulation, followed by simple concatenation to assemble the final program. Following \gls{sota} approaches~\cite{gao2025foundation, gao2025words}, we employ prompting techniques rather than fine-tuning: pre-trained \glspl{llm} already demonstrate strong performance and can be effectively adapted to domain-specific tasks via prompting techniques. Additionally, this approach preserves generalizability, enabling seamless deployment across different \gls{llm} models.

\subsubsection{Global Configuration Generator}\label{sec:global_config}
The Global Configuration Generator operates as a two-stage pipeline to detect and formalize scenario settings by using \textit{SettingsDetector} and \textit{HeaderGenerator} prompts. First, the \textit{SettingsDetector} analyzes the scenario description to detect global settings $G_{settings}$ using rule-based keyword matching patterns and example-based reasoning. The detector returns settings with confidence scores (only confidence $\geq$ 0.6 accepted, low-confidence or null detections use defaults: \texttt{Town05}, \texttt{ClearNoon}, \texttt{Lincoln\_mkz}). Second, the \textit{HeaderGenerator} transforms the detected settings into an executable Scenic header $G_{code} \gets \text{HeaderGenerator}(map, weather, models)$, covering CARLA map parameters, weather, and vehicles.

\subsubsection{Iterative Component Generator} \label{sec:iterative_component_generator}
\begin{figure}[t]
	\centering
	\includegraphics[width=1\linewidth]{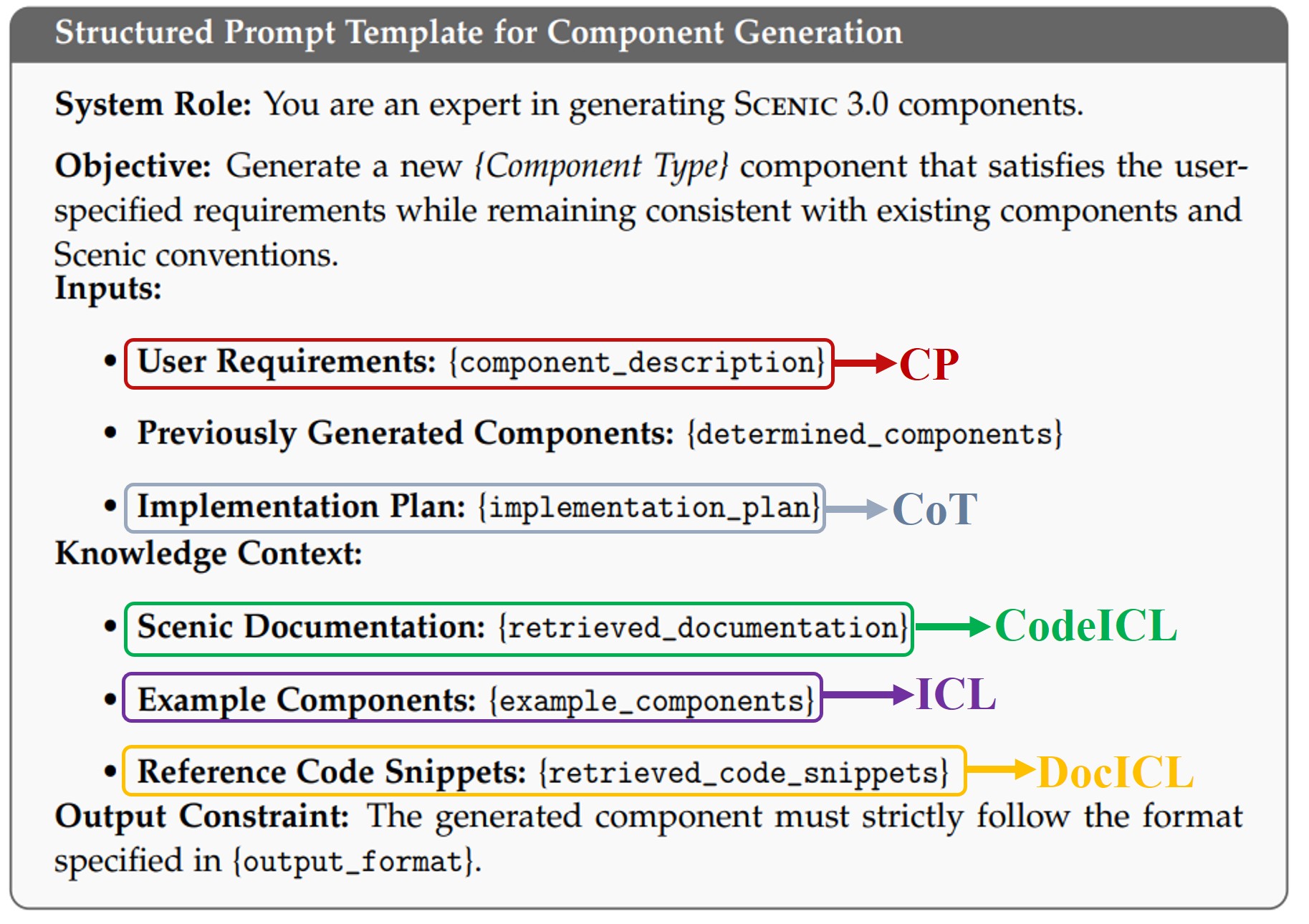}
    \vspace{-7mm}
	\caption{Prompting strategies for component generation.}
	\label{fig:prompting}
\end{figure}

Components are generated iteratively in dependency order using specialized generators: $\text{Generator}_R$ produces Spatial Relation code ($R_{code}$), $\text{Generator}_E$ produces Ego code ($E_{code}$), $\text{Generator}_O$ iteratively produces Object codes ($O_{1:N, code}$), and $\text{Generator}_T$ produces Restriction code ($T_{code}$). Finally, all components are concatenated to form the complete executable Scenic program $\mathcal{S}_{code} = \{G_{code}, R_{code}, E_{code}, O_{1:N, code}, T_{code}\}$. Each generator takes inputs including the logical description from related components in $\mathcal{S}_{int}$ and previously accumulated generated code context, ensuring compatibility between components. All generators share the same unified prompting strategy structure as shown in Fig.~\ref{fig:prompting}. 

\textbf{\gls{cp}:} Enhances the model's domain understanding by injecting domain-specific knowledge directly into the prompt, including syntax constraints, hierarchical type systems (e.g., for $\text{Generator}_R$: NetworkElement $\rightarrow$ LinearElement $\rightarrow$ \{Road, Lane\}), and available operators. This approach remains flexible, allowing additional domain knowledge to be incorporated.

\textbf{\gls{cot}:} Structures the generation task into explicit reasoning steps tailored per component (e.g., for Generator\_E: ``Understand requirements $\rightarrow$ Examine context $\rightarrow$ Select type $\rightarrow$ Define parameters $\rightarrow$ Define behavior $\rightarrow$ Instantiate $\rightarrow$ Validate''), enabling the model to complete each step incrementally, improving both understanding and generation accuracy.

\textbf{\gls{icl}:} Employs few-shot examples where the model learns from demonstration examples directly in the prompt context. Positive examples showcase correct implementation patterns and proper syntax, while negative examples present error-correction pairs. This contrastive learning approach anchors the model to concrete instances, reducing ambiguity and preventing common pitfalls.

\textbf{\gls{ragicl}:} Extends \gls{icl} with dynamic retrieval: the dual-retriever (Section~\ref{subsec:rag}) dynamically fetches the most relevant examples per query: \gls{codeicl} retrieves top-3 semantically similar code snippets, while \gls{docicl} retrieves relevant documentation chunks. Unlike fixed-example \gls{icl}, the dynamic retrieval adapts to each query, and the underlying database can be extended independently, making the framework flexible and generalizable to new domains.

Ablation studies on these prompting techniques are presented in Section~\ref{sec:ablation_studies}.

\section{Results \& Discussion}
\label{sec:results}
\subsection{Experimental Setup}
We conduct experiments using the 3D simulator CARLA 0.9.15 with the compatible \gls{dsl} Scenic 3.1.0. All experiments are performed on a Dell Alienware R15 equipped with an NVIDIA RTX 4090 GPU with 24GB VRAM.
\begin{figure*}[ht]
\centering
\includegraphics[width=0.90\linewidth]{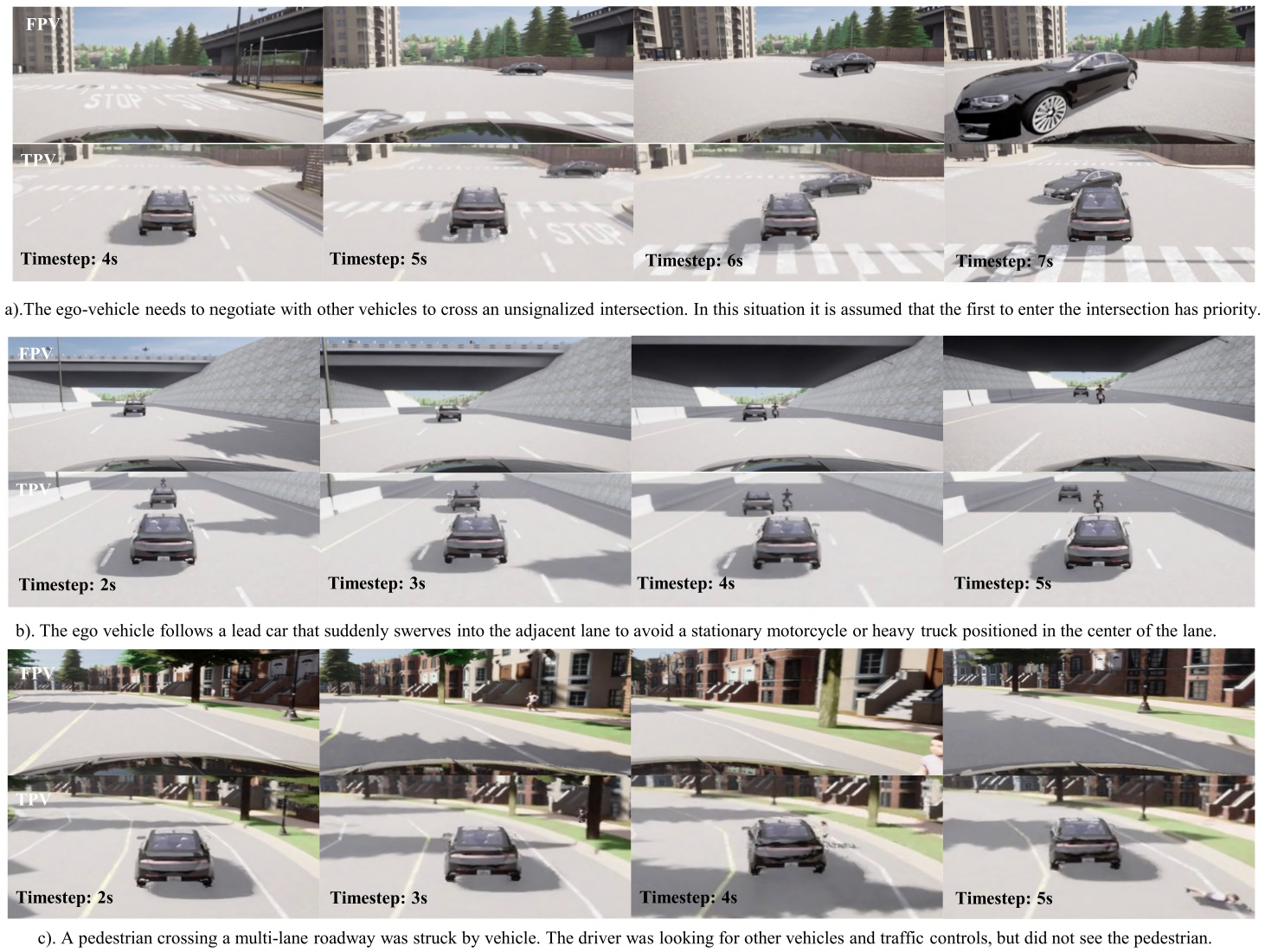}
\vspace{-2mm}
\caption{Example scenarios from three sources: a) CARLA leaderboard, b) UN R171, c) NHTSA Crash.}
\label{fig:qualitative_multiview_examples}
\end{figure*}
\subsubsection{Task Definition and Benchmark}
To evaluate our framework's performance, we propose a regulation-grounded benchmark with 123 scenario specifications drawn from multiple sources: the CARLA Leaderboard\footnote{\url{https://leaderboard.carla.org/}} (24 scenarios), NHTSA\footnote{\url{https://www.nhtsa.gov/}} crash and pre-crash datasets (47 scenarios), and United Nations Vehicle Regulations including UN R152, R157, and R171 (52 scenarios\footnote{\url{https://unece.org/transport/vehicle-regulations}}).  Table~\ref{tab:benchmark_distribution} presents the detailed distribution across these sources.
\begin{table}[t]
\centering
\caption{Distribution of scenarios in our benchmark.}
\label{tab:benchmark_distribution}
\begin{threeparttable}
\resizebox{\columnwidth}{!}{%
\renewcommand{\arraystretch}{1}
\setlength{\tabcolsep}{3pt}
\begin{tabular}{lcccccccc}
\toprule
\textbf{Source} & \textbf{Number} & \textbf{SCE} & \multicolumn{2}{c}{\textbf{VRU}} & \textbf{TC} & \textbf{Temporal} & \multicolumn{2}{c}{\textbf{DB}} \\
\cmidrule(lr){4-5} \cmidrule(lr){8-9}
& & & Number & Types & & & Vehicle & VRU \\
\midrule
CARLA Leaderboard & 24 & 24 & 3 & 2 & 3 & 11 & 13 & 2 \\
NHTSA Crash & 16 & 16 & 3 & 1 & 4 & 16 & 8 & 1 \\
NHTSA PreCrash & 31 & 31 & 4 & 2 & 4 & 27 & 13 & 2 \\
UN R152 & 4 & 3 & 1 & 2 & 1 & 2 & 7 & 1 \\
UN R157 & 12 & 12 & 6 & 2 & 0 & 12 & 6 & 4 \\
UN R171 & 36 & 35 & 9 & 3 & 0 & 35 & 8 & 3 \\
\midrule
\textbf{Total} & \textbf{123} & \textbf{121} & \textbf{26} & \textbf{3} & \textbf{12} & \textbf{103} & \textbf{20} & \textbf{10} \\
\bottomrule
\end{tabular}
}
\vspace{0.1em}
\begin{flushleft}
\scriptsize
SCE: Safety-critical Edge Cases, VRU: Vulnerable Road Users such as pedestrians and cyclists, TC: Traffic Control mechanisms such as signals and signs, Temporal: conditions affecting weather and time-of-day, DB: Dynamic Behaviors capturing motion patterns.
\end{flushleft}
\end{threeparttable}
\end{table}
\subsubsection{Evaluated Large Language Models}
Our evaluation includes leading proprietary and open-source \glspl{llm}. For the proprietary models, we evaluate across Qwen (Flash, Plus), DeepSeek-V3.2 in both Chat and Reasoner modes, Google Gemini-3 (Pro, Flash (no thinking mode)), specifically with Qwen3-Coder (Plus, Flash). Regarding the open-source models, we adopt Ollama\footnote{\url{https://ollama.com/}} with GPT-OSS 20B, Mistral-Small-3.2, Gemma-3-27B, Qwen3:30B and the code-specialized Qwen3-Coder:30B. 
\subsubsection{Evaluation Metrics}
To comprehensively evaluate our framework, we employ two categories of metrics:

\textbf{Framework Performance Metrics:} We assess the operational efficiency and reliability of the framework using: (i) \textit{Compilation Success Rate (CSR)}: percentage of generated Scenic programs that successfully execute in the simulator; (ii) \textit{\gls{rt}}: average generation time per scenario from description to code; (iii) \textit{\gls{tokens}}: average number of tokens consumed per scenario.  

\textbf{Scenario Generation Accuracy:} For scenarios that successfully compile, we evaluate how accurately the generated scenario reflects the original description. Since our output is executable code, language quality metrics (e.g., BLEU, ROUGE-L) are not applicable. Instead, we adopt a human evaluation structured by a layers model~\cite{scholtes20216}, which decomposes a driving scenario into semantically distinct layers. Human evaluator compares each layer of the generated scenario against the scenario description and assigns a correctness score. Specifically, the evaluated layers are: (i) \textit{Road}: geometry and topology; (ii) \textit{Traffic Infrastructure}: traffic light and signs; (iii) \textit{Temporal Modifications}: construction zones; (iv) \textit{Objects}: Static and Dynamic Objects; (v) \textit{Environment}: weather and time-of-day conditions; (vi) \textit{\gls{sq}}: the average alignment score across layers. Additionally, we report the \textit{Framework Accuracy (FA)}, computed as \gls{fa} = \gls{csr} × \gls{sq}, which represents the overall end-to-end accuracy.

\subsection{Experiments and Results}
Fig.~\ref{fig:qualitative_multiview_examples} shows CARLA-rendered qualitative examples from three benchmark sources: (a) an ego vehicle negotiating an unsignalized intersection (CARLA Leaderboard), (b) an ego vehicle following a lead car that swerves to avoid a stationary obstacle (UN R171), and (c) a pedestrian crossing a multi-lane roadway undetected (NHTSA Crash). We further conduct comprehensive ablation studies and cross-model comparisons on the proposed benchmark.

\subsubsection{Ablation Study}\label{sec:ablation_studies}
As discussed in Section~\ref{sec:iterative_component_generator}, we combine \gls{cp}, \gls{cot}, \gls{icl}, and \gls{ragicl} to inject domain-specific knowledge and guide \gls{llm} generation. We systematically evaluate 12 configurations (C1-C12 in Table~\ref{tab:ablation_results}) organized into four tiers using Gemini 3 Flash: (i) \textit{Zero/Few-shot Tier} (C1-C3): baseline and \gls{icl}/\gls{codeicl} without \gls{cp}; (ii) \textit{Contextual Tier} (C4-C7): \gls{cp} as foundation with various combinations; (iii) \textit{Thinking Tier} (C8-C11): \gls{cp}+\gls{cot} with various combinations; (iv) \textit{Extended} (C12): all techniques including \gls{docicl}. Results are summarized in Table~\ref{tab:ablation_results}.
\begin{table*}[t]
    \centering
    \caption{Ablation study results across all prompting technique combinations (based on Gemini-3-Flash model).}
    \label{tab:ablation_results}
    \small
    \begin{threeparttable}
    \resizebox{\textwidth}{!}{
    \renewcommand{\arraystretch}{0.85}
    \begin{tabular}{lccccccccccccccc}
    \toprule
    \multirow{2}{*}{\textbf{ID}} & \multirow{2}{*}{\textbf{CP}} & \multirow{2}{*}{\textbf{CoT}} & \multirow{2}{*}{\textbf{ICL}} & \multicolumn{2}{c}{\textbf{RAG-ICL}} & \multirow{2}{*}{\colorbox{blue!15}{\textbf{CSR (\%)}$\uparrow$}} & \multirow{2}{*}{\textbf{RT (s)}$\downarrow$} & \multirow{2}{*}{\textbf{Tokens}$\downarrow$} & \multicolumn{6}{c}{\textbf{Scenario Alignment(\%)}$\uparrow$} & \multirow{2}{*}{\colorbox{blue!15}{\textbf{FA(\%)}$\uparrow$}}\\
    \cmidrule(lr){5-6} \cmidrule(lr){10-15}
    & & & & \textbf{CodeICL} & \textbf{DocICL} & & & & \textbf{RL}$\uparrow$ & \textbf{TI}$\uparrow$ & \textbf{TM}$\uparrow$ & \textbf{DO}$\uparrow$ & \textbf{EN}$\uparrow$ & \cellcolor{blue!12}\textbf{SQ}$\uparrow$ &  \\
    \midrule
    \rowcolor[gray]{0.95} \multicolumn{16}{l}{\textit{Zero/Few-shot Tier}} \\
    C1$^*$ & -- & -- & -- & -- & -- & 0.00 & 170.63 & 32572 & 0.00 & 0.00 & 0.00 & 0.00 & 0.00 & 0.00 & 0.00 \\
    C2 & -- & -- & \checkmark & -- & -- & 4.06 & 102.58 & 22889 & 100.00 & 100.00 & -- & 66.67 & 100.00 & 88.89 & 3.61 \\
    C3 & -- & -- & -- & \checkmark & -- & 8.10 & 157.75 & 27796 & 64.58 & 75.00 & 100.00 & 41.67 & 75.00 & 54.86 & 4.44 \\
    \midrule
    \rowcolor[gray]{0.95} \multicolumn{16}{l}{\textit{Contextual Tier}} \\
    C4 & \checkmark & -- & -- & -- & -- & 12.20 & 435.59 & 43657 & 82.29 & 100.00 & 100.00 & 52.08 & 100.00 & 69.79 & 8.51 \\
    C5 & \checkmark & -- & \checkmark & -- & -- & 47.15 & 196.60 & 39702 & 88.10 & 83.33 & 76.67 & 43.12 & 100.00 & 69.18 & 32.62 \\
    C6 & \checkmark & -- & -- & \checkmark & -- & 58.54 & 318.92 & 47085 & 84.43 & 90.00 & 75.00 & 45.61 & 97.62 & 69.17 & 40.49 \\
    C7 & \checkmark & -- & \checkmark & \checkmark & -- & 60.10 & 172.27 & 48606 & 88.22 & 78.57 & 85.00 & 54.89 & 98.00 & 74.93 & 45.03 \\
    \midrule
    \rowcolor[gray]{0.95} \multicolumn{16}{l}{\textit{Thinking Tier}} \\
    C8 & \checkmark & \checkmark & -- & -- & -- & 13.01 & 624.12 & 55153 & 84.31 & 100.00 & -- & 41.18 & 100.00 & 65.52 & 8.52 \\
    C9 & \checkmark & \checkmark & \checkmark & -- & -- & 54.47 & 520.47 & 52689 & 87.81 & 92.31 & 100.00 & 49.50 & 97.22 & 71.77 & 39.09 \\
    C10 & \checkmark & \checkmark & -- & \checkmark & -- & 60.98 & 300.47 & 46005 & 84.46 & 77.78 & 83.33 & 51.58 & 100.00 & 70.70 & 43.11 \\
    \rowcolor[HTML]{D4EDDA} C11 & \checkmark & \checkmark & \checkmark & \checkmark & -- & \textbf{76.42} & 222.11 & 52339 & 90.65 & 75.00 & 83.33 & 58.16 & 96.00 & 76.12 & \textbf{58.17} \\
    \midrule
    \rowcolor[gray]{0.95} \multicolumn{16}{l}{\textit{Extended Configuration}} \\
    C12 & \checkmark & \checkmark & \checkmark & \checkmark & \checkmark & 56.90 & 563.36 & 70225 & 86.62 & 78.57 & 75.00 & 54.93 & 100.00 & 73.30 & 41.71 \\
    \bottomrule
    \end{tabular}}
    \begin{flushleft}
    \scriptsize
    \vspace{0.2em}
    Arrows indicate if higher ($\uparrow$) or lower ($\downarrow$) values are better. $^*$C1 is a zero-shot baseline model. 
    CP: Contextual Prompting, CoT: Chain-of-Thought, ICL: In-Context Learning, RAG-ICL: Retrieval-Augmented Generation with In-Context Learning, including Code and Doc. 
    CSR: Compilation Success Rate (\%), RT: Response Time (s), Tokens: Token usage. Scenario Alignment Layers: RL: Road, TI: Traffic Infrastructure, TM: Temporal Modifications, DO: Dynamic Objects, EN: Environment. SQ: Scenario Quality, FA: Framework Accuracy.
    \end{flushleft}
    \end{threeparttable}
    \end{table*}

\begin{table*}[t]
\centering
\caption{Comparison across LLM backbones and SOTA methods evaluated on the proposed 123-scenario benchmark.}
\label{tab:cross_model_comparison}
\small
\begin{threeparttable}
\resizebox{\textwidth}{!}{
\renewcommand{\arraystretch}{0.85}
\begin{tabular}{lcccccccccc}
\toprule
\multirow{2}{*}{\textbf{Model Variant}} & \multirow{2}{*}{\colorbox{blue!15}{\textbf{CSR (\%)}$\uparrow$}} & \multirow{2}{*}{\textbf{RT (s)}$\downarrow$} & \multirow{2}{*}{\textbf{Tokens}$\downarrow$} & \multicolumn{6}{c}{\textbf{Scenario Alignment(\%)}$\uparrow$} & \multirow{2}{*}{\colorbox{blue!15}{\textbf{FA(\%)}$\uparrow$}}\\
\cmidrule(lr){5-10}
& & & & \textbf{RL}$\uparrow$ & \textbf{TI}$\uparrow$ & \textbf{TM}$\uparrow$ & \textbf{DO}$\uparrow$ & \textbf{EN}$\uparrow$ & \cellcolor{blue!12}\textbf{SQ}$\uparrow$ &  \\
\midrule
\rowcolor[gray]{0.95} \multicolumn{11}{l}{\textit{Open-Source Models}} \\
Qwen3-Coder:30B & 0.00 & 62.95 & 21098 & 0.00 & 0.00 & 0.00 & 0.00 & 0.00 & 0.00 & 0.00 \\
Qwen3:30B & 0.00 & 135.23 & 30659 & 0.00 & 0.00 & 0.00 & 0.00 & 0.00 & 0.00 & 0.00 \\
Gemma3:27B & 0.00 & 121.88 & 35705 & 0.00 & 0.00 & 0.00 & 0.00 & 0.00 & 0.00 & 0.00 \\
GPT-OSS:20B & 0.82 & 96.69 & 33199 & 100.00 & 100.00 & -- & 100.00 & -- & 100.00 & 0.82 \\
Mistral-Small3.2:24B & 1.62 & 125.74 & 32129 & 100.00 & 33.33 & -- & 11.11 & 100.00 & 58.33 & 0.94 \\
\midrule
\rowcolor[gray]{0.95} \multicolumn{11}{l}{\textit{Proprietary Models}} \\
Qwen-Flash & 1.63 & 15.00 & 22081 & 83.33 & 100.00 & -- & 22.22 & -- & 62.04 & 1.01 \\
Qwen-Plus & 5.70 & 14.05 & 21937 & 100.00 & 100.00 & -- & 66.67 & 100.00 & 87.30 & 4.98 \\
Qwen3-Coder-Plus & 10.57 & 19.67 & 22072 & 85.71 & 80.00 & -- & 34.52 & 91.67 & 65.28 & 6.90 \\
Qwen3-Coder-Flash & 15.44 & 13.85 & 21980 & 97.37 & 60.00 & 33.33 & 39.47 & 100.00 & 70.61 & 10.90 \\
DeepSeek-V3.2 (Chat) & 12.20 & 28.04 & 22676 & 85.09 & 70.00 & 0.00 & 45.61 & 100.00 & 72.37 & 8.83 \\
DeepSeek-V3.2 (Reasoner) & 14.63 & 460.62 & 35290 & 93.75 & 60.00 & 100.00 & 56.25 & 100.00 & 78.30 & 11.46 \\
Gemini-3-Pro & 60.16 & 165.39 & 47928 & 92.22 & 61.54 & 85.71 & 53.33 & 100.00 & 73.50 & 44.22 \\
\rowcolor[HTML]{D4EDDA}\textbf{Gemini-3-Flash} & \textbf{76.42} & 222.11 & 52339 & 90.65 & 75.00 & 83.33 & 58.16 & 96.00 & 76.12 & \textbf{58.17} \\
\midrule
\rowcolor[gray]{0.95} \multicolumn{11}{l}{\textit{SOTA (Gemini-3 Flash)}} \\
ChatScene~\cite{zhang2024chatscene} (Retrieval Assemble) & 30.08 & 10.02 & 4476 & 65.03 & 37.88 & 0 & 7.52 & 46.67 & 36.68 & 11.03 \\
NL2Scenic~\cite{bauerfeind2025david} (Retrieval Generation) & 16.26 & 43.44 & 20444 & 96.15 & 100.00 & 66.67 & 33.33 & 100.00 & 66.77 & 10.86 \\
\bottomrule
\end{tabular}}
\end{threeparttable}
\begin{flushleft}
\vspace{-1mm}
\scriptsize
Arrows indicate if higher ($\uparrow$) or lower ($\downarrow$) values are better. 
% CSR: Compilation Success Rate, RT: Response Time (s). Scenario Alignment Layers: RL: Road, TI: Traffic Infrastructure, TM: Temporal Modifications, DO: Dynamic Objects, EN: Environment. SQ: Scenario Quality, FA: Framework Accuracy.
\end{flushleft}
\end{table*}
The results demonstrate clear performance improvements as more advanced techniques are incorporated. Starting from the zero-shot baseline (C1) with \gls{csr} and \gls{fa} of 0\%, adding \gls{cp} alone (C4) increases \gls{csr} to 12.20\% and \gls{fa} to 8.51\%. Combining \gls{cp} with \gls{icl} (C5) significantly boosts performance \gls{csr} of 47.15\% and \gls{fa} of 32.62\%. Further adding \gls{cot} to create the thinking tier (C9) achieves \gls{csr} of 54.47\% and \gls{fa} of 39.09\%. The best performance is achieved by C11, which incorporates \gls{cp}, \gls{cot}, \gls{icl}, and \gls{codeicl}, achieving a \gls{csr} of 76.42\%, \gls{sq} of 76.12\%, and \gls{fa} of 58.17\% with a response time of 222.11 seconds. Notably, adding \gls{docicl} (C12) does not improve performance, as the Scenic documentation is better suited for our interactive chats to understand code concepts, rather than providing practical guidance for code generation, which incurs higher time and token costs. A crucial observation is that providing more code examples through \gls{codeicl} significantly improves generation efficiency: this emerges by comparing C7 (\gls{cp}+\gls{icl}+\gls{codeicl}, \gls{rt}=172.27s) with C5 (\gls{cp}+\gls{icl}, \gls{rt}=196.60s), and C11 (\gls{cp}+\gls{cot}+\gls{icl}+\gls{codeicl}, \gls{rt}=222.11s) with C9 (\gls{cp}+\gls{cot}+\gls{icl}, \gls{rt}=520.47s), as code snippets help the model converge faster to correct solutions.

\subsubsection{Comparison across Models and \gls{sota} Methods}
As shown in Table~\ref{tab:cross_model_comparison}, we evaluate our Chat2Scenic framework (C11 configuration) across leading proprietary, open-source \gls{llm} and \gls{sota} methods. The Gemini-3 family models outperform other proprietary models, likely due to their powerful capabilities and support for long context windows. Notably, Gemini-3-Flash achieves the best overall performance with \gls{csr} of 76.42\% and \gls{fa} of 58.17\%, significantly outperforming Gemini-3-Pro (\gls{csr} 60.16\%, \gls{fa} 44.22\%). This aligns with findings from~\cite{gao2025nurisk} that internal reasoning of thinking models can be hindered by external prompting strategies, whereas instruction-optimized models like Flash are better suited for structured prompting techniques. In contrast, open-source models perform poorly: most achieve near-zero CSR (0\%--1.62\%), primarily due to their smaller scale, which limits general capabilities, less extensive pre-training data compared to proprietary models, and difficulty following complex prompting strategies with multiple interleaved techniques.

Furthermore, based on the same Gemini-3-Flash model, we compare our framework against \gls{sota} methods, specifically ChatScene~\cite{zhang2024chatscene} (Retrieval Assemble) and NL2Scenic~\cite{bauerfeind2025david} (Retrieval Generation), which provide publicly available implementations. ChatScene achieves better performance (\gls{csr} 30.08\%) than NL2Scenic, as it leverages a pre-built code database from CARLA Leaderboard scenarios for retrieval-based assembly. In contrast, NL2Scenic's direct retrieval of complete Scenic scripts struggles with complex scenarios despite advanced prompting, as full-script generation lacks the flexibility and stability needed for diverse scenario specifications. Our framework significantly outperforms both methods, achieving a \gls{csr} of 76.42\%, demonstrating that iterative component-wise generation with diverse prompting strategies improves the effectiveness and robustness of scenario generation. While our approach requires longer response time (\gls{rt} of 222.11 seconds) due to iterative generation, this is an acceptable compromise for offline scenario generation, where accuracy and reliability are prioritized over generation speed.

\section{Conclusion and Future Work}
\label{sec:conclusion}
We presented Chat2Scenic, the first iterative retrieval-augmented framework for \gls{ads} scenario generation. To enable systematic evaluation, we introduced a regulation-grounded benchmark comprising 123 scenarios derived from multiple sources, along with comprehensive evaluation metrics. Through ablation studies and cross-model comparisons, we demonstrated that advanced prompting techniques (\gls{cp}+\gls{cot}+\gls{icl}+\gls{codeicl}) achieve \gls{csr} of 76.42\%, \gls{sq} of 76.12\%, and \gls{fa} of 58.17\% (with Gemini-3-Flash), significantly outperforming existing retrieval-assemble and full-script generation approaches.
Future work will extend the framework with multimodal inputs (e.g., images and sketches) and simulation-in-the-loop feedback to enhance scenario specification and generation accuracy.
\section*{ACKNOWLEDGMENT}
The manuscript was initially drafted by the authors, with AI tools used to improve grammar and clarity.

\bibliographystyle{IEEEtran}
\bstctlcite{IEEEexample:BSTcontrol}
\bibliography{literature}

%%%%%%%%%%%%%%%%%%%%%%%%%%%%%%%%%%%%%%%%%%%%%%%%%%%%%%%%%%%%%%%%%%%%%%%%%%%%%%%%%%%%%%%%%%%%%%%%%%%%%%%%%%%%%%%%
    
\end{document}